\documentclass{article}
\pdfoutput=1
\usepackage{ijcai19}

\usepackage{natbib}

\usepackage{times}
\usepackage{soul}
\usepackage{url}
\usepackage[hidelinks]{hyperref}
\usepackage[utf8]{inputenc}
\usepackage[labelfont={bf},small]{caption}
\usepackage{enumitem}
\usepackage[usenames,dvipsnames,svgnames,table]{xcolor}
\usepackage{float}
\usepackage{graphicx}
\usepackage{amsmath}
\usepackage{booktabs}
\usepackage{tcolorbox}
\usepackage{latexsym} 
\usepackage{dingbat}
\usepackage{balance}
\usepackage{multicol, multirow}
\usepackage{color}
\usepackage{adjustbox}
\usepackage{expex}
\usepackage{textcomp}
\usepackage{siunitx}
\usepackage{comment}
\usepackage{soul}

\usepackage{tabularx, ragged2e, array, arydshln}
\usepackage[normalem]{ulem}

\usepackage{todonotes}

\let\oldhdashline\hdashline
\renewcommand{\hdashline}{\noalign{\vskip 0.5ex}\oldhdashline\noalign{\vskip 0.5ex}}

\defcitealias{ufal}{ÚFAL, 2017}

\begin{document}

\pagenumbering{arabic}

\newcommand{\rtodo}[1]{\textbf{\textcolor{red}{TODO\ #1}}}
\newcommand{\idea}[1]{\footnotesize\noindent\textit{\textcolor{blue}{#1\\}}}


\title{EriBERTa: \\ A Bilingual Pre-Trained Language Model for \\Clinical Natural Language Processing}

\author{
Iker de la Iglesia\and
Aitziber Atutxa \and
Koldo Gojenola\and
Ander Barrena \\[0.5em]
\affiliations
HiTZ Basque Center for Language Technology \\[0.2em] University of the Basque Country (UPV/EHU), Spain
\emails
\{first-name\}.\{last-name\}@ehu.eus
}

\maketitle

\begin{abstract}
The utilization of clinical reports for various secondary purposes, including health research and treatment monitoring, is crucial for enhancing patient care. Natural Language Processing (NLP) tools have emerged as valuable assets for extracting and processing relevant information from these reports. However, the availability of specialized language models for the clinical domain in Spanish has been limited.

In this paper, we introduce EriBERTa, a bilingual domain-specific language model pre-trained on extensive medical and clinical corpora. We demonstrate that EriBERTa outperforms previous Spanish language models in the clinical domain, showcasing its superior capabilities in understanding medical texts and extracting meaningful information. Moreover, EriBERTa exhibits promising transfer learning abilities, allowing for knowledge transfer from one language to another. This aspect is particularly beneficial given the scarcity of Spanish clinical data.
\end{abstract}

\section{Introduction}
\label{sec:introduction}

Extracting and processing relevant information from clinical reports pose significant challenges due to their unstructured nature and the domain-specific language used in healthcare. Natural Language Processing (NLP) techniques have emerged as powerful tools for tackling these challenges and unlocking the potential of clinical data.

The transformer architecture, introduced by \citet{vaswani2017attention}, has revolutionized NLP by enabling the training of deep neural networks capable of capturing contextual relationships between words and producing state-of-the-art results across various tasks. This architecture has been successfully applied to numerous domains, including healthcare.

In the medical and clinical domains, the development of domain-specific language models has further advanced the analysis of clinical data. Models such as SciBERT \citep{beltagy2019scibert}, BioBERT \citep{lee2020biobert}, and BioALBERT \citep{naseem2021bioalbert} have been specifically designed and trained on biomedical literature to capture the intricacies of medical terminology and context. These models have demonstrated remarkable performance in various biomedical NLP tasks, including named entity recognition, relation extraction, and document classification in English. 
However, progress in the Spanish language has been limited, primarily due to the difficulty of obtaining clinical data and the resulting scarcity of available Spanish corpora. Nevertheless, models such as roberta-base-biomedical-es \citep{carrino2021biomedical} and bsc-bio-ehr-es \citep{carrino-etal-2022-bsc-bio-ehr-es} have been developed and shown promising performances in Spanish.

To address the challenges posed by limited data availability in the Spanish language, we present EriBERTa, a bilingual domain-specific language model tailored specifically for the medical and clinical domains. EriBERTa is pre-trained on medical and clinical corpora in both English and Spanish, enabling it to effectively capture the nuances of medical terminology and understand the context of clinical narratives. By bridging the gap between English and Spanish models, EriBERTa opens up new opportunities for NLP applications in Spanish-speaking healthcare settings.

In this paper, firstly, we provide a detailed analysis of EriBERTa's architecture. Secondly, we outline the training methodology employed to pre-train EriBERTa on a large-scale medical and clinical corpus, enabling it to learn the intricacies of medical language. Lastly, we conduct comprehensive evaluations of EriBERTa on benchmark datasets to demonstrate its superiority in both monolingual and cross-lingual settings in Spanish, while having competitive performance in English datasets. 
The experimental findings highlight the potential of EriBERTa to enhance the extraction of meaningful insights from clinical data and contribute to advancements in healthcare research and practice.



\section{Pre-Training EriBERTa}
\label{sec:materials}

The EriBERTa model is a pre-trained transformer-based language model designed to enhance the performance of natural language understanding tasks in the medical-clinical domain for the Spanish language. It is based on the RoBERTa architecture, which is a popular pre-trained model and a variant of BERT \citep{devlin2018bert}. EriBERTa uses the same tokenizer and pre-training methods as RoBERTa, as described in the work of \citet{roberta}. This ensures that EriBERTa has similar performance characteristics to RoBERTa but with the additional advantage of being fine-tuned on English and Spanish medical-clinical data. We trained three variations of EriBERTa: one using public corpora, another using private hospital EHR documents, and a Longformer version based on the one trained with private hospital data.

In this section, we will delve into the details of the  pre-training and evaluation procedures utilized in our research.
Specifically, we will begin by describing the corpora used for the Masked Language Modeling (MLM) pre-training and will highlight the differences between the corpora used for both EriBERTa versions.
We will then proceed to explain the specific pre-training procedures for each EriBERTa version. 
Finally, we will present the datasets used for evaluation, and provide a detailed account of the evaluation setup for each dataset.

\subsection{Corpora}
\label{sec:materials:corpora}

To pre-train the EriBERTa model, we used a combination of publicly available and private medical-clinical corpora in Spanish and English. The selection of corpora was based on their relevance to the medical domain and the availability of high-quality data. However, obtaining medical and clinical corpora can be challenging, especially in the Spanish language.

One of the main difficulties in obtaining medical and clinical corpora is the sensitive nature of the data. Medical records contain highly personal and confidential information, which makes it challenging to collect and share this type of data for research purposes. 
These challenges are compounded when it comes to the Spanish language. Compared to English, there are fewer medical and clinical corpora available in Spanish, which can make it challenging to obtain sufficient data for pre-training language models. Furthermore, the quality and quantity of available Spanish corpora can be highly variable, which can impact the performance of pre-trained models.

Despite the challenges of gathering medical and clinical corpora, we have successfully assembled and curated a diverse collection of relevant corpora in both Spanish and English to pre-train the EriBERTa model. \autoref{tab:pretrain-corpus} provides more detailed information on the language and size of each corpus used for pretraining.

\begin{table}[htb]
\centering
\begin{tabular}{@{}llr@{}}
\toprule
\textbf{Lang} & \multicolumn{1}{c}{\textbf{Source}} & \textbf{No. Words} \\ \midrule
\multicolumn{3}{c}{\cellcolor[HTML]{EFEFEF}\textit{Medical Corpus}}                                                \\
\multirow{3}{*}{ENG}              & EMEA                                & 12M               \\
                                  & PubMed Abstracts                    & 968.4M             \\
                                  & Clinical Trials                     & 127.4M             \\\hdashline
\multirow{7}{*}{ES}               & EMEA                                & 13.6M              \\
                                  & PubMed                              & 8.4M               \\
                                  & SNOMED-CT                           & 7.2M              \\
                                  & SPACCC                              & 350k              \\
                                  & UFAL                                & 10.5M              \\
                                  & Wikipedia (Med)                     & 5.2M               \\
                                  & Medical Crawler                     & 918M               \\\midrule
\multicolumn{3}{c}{\cellcolor[HTML]{EFEFEF}\textit{Clinical Corpus (EHR)}}                         \\
ENG                               & MIMIC-III                           & 206M               \\\hdashline
ES                                & Private Hospital Documents          & 222M               \\ \bottomrule
\end{tabular}
\caption{Pre-training corpora used classified by document type and language.}
\label{tab:pretrain-corpus}
\end{table}

The corpora we used for pretraining EriBERTa are:

\begin{itemize}
    \item \textbf{MIMIC-III} \citep{mimiciii}: English database on ICU stays of over 40,000 patients between 2001 and 2012. It contains information such as vital sign measurements, laboratory test results, procedures, medications, caregiver notes, and mortality, among others.
    
    \item \textbf{EMEA} \citep{tiedemann2012emea}: A parallel corpus in English and Spanish consisting of documents from the European Medicines Agency.
    
    \item \textbf{ClinicalTrials}\footnote{\url{https://clinicaltrials.gov}}: Set of English documents on clinical studies carried out worldwide.
    
    \item \textbf{PubMed}: Contains abstracts and full texts of biomedical literature from multiple NLM literary sources including MEDLINE\footnote{\label{medline}\url{https://www.nlm.nih.gov/medline/medline_overview.html}}, PubMed Central\footnote{\label{pubmedCentral}\url{https://www.ncbi.nlm.nih.gov/pmc/about/intro/}}, and Bookshelf\footnote{\label{bookshelf}\url{https://www.ncbi.nlm.nih.gov/books/}}. We used abstracts for English and abstracts and full texts for Spanish.
    
    \item \textbf{SNOMED-CT} \citep{snomed}: Standardized and multilingual medical vocabulary consisting of more than 300,000 medical concepts, including categories such as body parts, clinical findings, and pharmaceutical/biological products, among others. For this work, the descriptions in Spanish associated with each term were used.
    
    \item \textbf{SPACCC} \citep{spaccc}: Spanish corpus created after collecting 1,000 clinical cases from SciELO\footnote{\url{https://scielo.isciii.es/scielo.php}}, and categorizing them based on structure and content into those that were similar to real clinical texts and those that were not suitable for this task.
    
    \item \textbf{UFAL} \citepalias{ufal}: Multilingual medical corpus composed of parallel corpora that have been collected over various projects.
    
    \item \textbf{Wikipedia Med}: A Spanish corpus composed of entries collected from Wikipedia, filtered by scope, and cleaned.
    
    \item \textbf{Private Clinical Documents}: Set of Spanish clinical narrative from health centers. It is used exclusively on the private version of EriBERTa.
    
    \item \textbf{Medical Crawler} \citep{carrino2021medcrawler}: A corpus comprising over 3,000 URLs related to Spanish biomedical and health domains, collected through web crawling. It is used exclusively on the public version of EriBERTa.
\end{itemize}

It is important to note that the main difference between the private and public versions of EriBERTa is the type of clinical documents used for pretraining. 
The private version of EriBERTa was trained using the Private Clinical Documents corpus, but not the Medical Crawler corpus. In contrast, the public version of EriBERTa was trained using the Medical Crawler corpus, but not the Private Clinical Documents corpus. 
These differences in pretraining corpora may affect the performance of the models on different tasks, particularly those that involve hospital clinical documents.

\subsubsection{Balancing the Corpus}

As shown in \autoref{tab:pretrain-corpus}, the amount of resources in Spanish and English for the private version of EriBERTa is imbalanced (1,313M words in English and 267M in Spanish). This imbalance could result in English having too much weight when generating the tokenizer and pretraining the model, leading to poor performance in Spanish, as reported in \citet{conneau2019xmlr}. Notably, the private EriBERTa version that uses the crawler does not suffer from this imbalance and thus does not require balancing.

To address this issue, we employed the formula \eqref{eqn:balanceo_corpus} proposed in \citet{Conneau2019CrosslingualLM}. Here, $n_i$ and $p_i$ represent the number of words and the frequency of occurrence of the language $i$, respectively, and $q_i$ denotes the probability that a word in language $i$ is sampled according to the multinomial distribution. $\alpha$ is a parameter that controls the language sampling rate, with lower values reducing the sampling probability of the most represented languages and increasing the probability of those with scarce resources. Based on the studies described in \citet{conneau2019xmlr} for multilingual models with few resources for some languages, we decided to balance the corpus using a parameter of $\alpha = 0.3$.

\begin{equation}
\label{eqn:balanceo_corpus}
{q_i}{i=1...N} \ \ \textrm{;} \ \ q_i = \frac{p^\alpha_i}{\sum{j=1}^{N}{p^\alpha_j}} \ \ \textrm{where} \ \ p_i= \frac{n_i}{\sum_{j=1}^{N}{n_j}}
\end{equation}

In the corpus balancing calculations, we decided to omit the section related to the EHR files because it was already balanced: 206M in English and 222M in Spanish. Without considering these files and applying the formulas in \eqref{eqn:balanceo_corpus}, we obtain:

\begin{equation}
\label{eqn:balanceo_corpus_results}
p_{es} = 0.04 \ \ \textrm{;} \ \
p_{eng} = 0.96 \ \ \textrm{;} \ \
q_{es} = 0.277 \ \ \textrm{;} \ \
q_{eng} = 0.723
\end{equation}

Therefore, in order for the probabilities $q_{es}$ and $q_{eng}$ obtained in \eqref{eqn:balanceo_corpus_results} to hold, we must multiply the Spanish texts by 9.5 times, from 45.4M tokens to 438M tokens.

\subsection{Model Pre-Training}
\label{sec:materials:pretraining}


In this section, we detail the pre-training process of our EriBERTa models, which are based on the RoBERTa-base \citep{roberta} and Longformer \citep{beltagy2020longformer} architectures. We first describe the tokenizer we used, which is the same as RoBERTa, followed by the vocabulary and pretraining details. Finally, we will delve into the Longformer variant pre-training details.

Our tokenizer is based on Byte-Pair Encoding (BPE), introduced by \citet{sennrich-etal-2016-neural}, which is a data compression technique used to represent a large set of symbols or words using a smaller vocabulary converting them to subwords, allowing the tokenizer to handle rare or out-of-vocabulary words. 
We defined a cased vocabulary with a size of 64,000 that was chosen to accommodate the bilingual nature of the models and ensure sufficient coverage of both languages \citep{conneau2019xmlr}.

For pre-training, we used the Masked Language Modeling (MLM) objective for all model variants, which involves randomly masking tokens in the input sequence and training the model to predict the original tokens based on the surrounding context \citep{devlin2018bert}.
To generate the pre-training data, we concatenated all the raw text from the corpora and then split the concatenated text into segments of the maximum input length allowed by each model.  
We trained the RoBERTa-based EriBERTa models from scratch for 125k steps, with checkpoints saved every 2.5k steps. The model parameters and training hyperparameters are defined in \autoref{tab:materials:pretraining:hiperparams}. 

\begin{table}[htb]
\centering

\begin{adjustbox}{max width=\linewidth}
\begin{tabular}{@{}lclc@{}}
\toprule
\multicolumn{4}{c}{\textbf{Training Hyperparameters}}                                                                                   \\ \midrule
\textit{Number of Layers}      & 12       & \textit{Batch Size (tokens)} & 2,083,840          \\
\textit{Hidden Size}           & 768      & \textit{Weight Decay}        & 0.0                \\
\textit{FFN inner hidden size} & 3,072     & \textit{Max Steps}           & 125k               \\
\textit{Attention heads}       & 12       & \textit{Learning Rate Decay} & Linear with warmup \\
\textit{Dropout}               & 0.1      & \textit{Adam $\epsilon$}     & 1e-08              \\
\textit{Attention Dropout}     & 0.1      & \textit{Adam $\beta_{1}$}    & 0.9                \\
\textit{Warmup Steps}          & 7.5k     & \textit{Adam $\beta_{2}$}    & 0.99               \\
\textit{Peak Learning Rate}    & 2.683e-4 & \textit{Gradient Clipping}   & 1                  \\ \bottomrule
\end{tabular}
\end{adjustbox}

\caption{Parameter and hyperparameter details for EriBERTa models.}
\label{tab:materials:pretraining:hiperparams}
\end{table}

\subsubsection{Pre-Training EriBERTa-Longformer}

Medical and clinical documents often pose a challenge for most transformer-based models due to their length which frequently exceeds the token limit of 512 or approximately 320 words. Various approaches have been proposed to address this issue, including text summarization, truncation, or splitting by paragraphs \citep{sun2019bert_long_documents_text_classification}. However, these methods may not be suitable for tasks where global context is crucial, and the text cannot be altered, such as clinical section detection.

To overcome this challenge, we introduce a Longformer variant of {EriBERTa}, based on the Longformer architecture \citep{beltagy2020longformer}. The Longformer extends the input sequence length by using a sliding window mechanism up to 4,096 tokens, enabling access to longer contexts in medical documents. The Longformer implementation also includes a global attention mechanism that attends to all tokens in the input sequence, which can capture dependencies beyond the local context. It is important to note that the global attention mechanism must be manually configured for each task and is not pre-trained.

The Longformer version of EriBERTa was initialized with the pre-trained weights of the EriBERTa model trained with private clinical documents using the process defined by \citet{beltagy2020longformer}. We further pre-trained the model using the MLM objective with the same corpus used for the original EriBERTa model. This step allowed the model to adapt to the new attention system using the same hyperparameters described in \citet{beltagy2020longformer}, as represented in \autoref{tab:materials:pretraining:hiperparams-longformer}.

\begin{table}[htb]
\centering
\begin{tabular}{@{}lc@{}}
\toprule
\multicolumn{2}{c}{\textbf{Training Hyperparameters}} \\ \midrule
\textit{Max Steps}             & 65k                  \\
\textit{Batch Size (tokens)}   & 262.144              \\
\textit{Warmup Steps}          & 500                  \\
\textit{Peak Learning Rate}    & 3e-5                 \\
\textit{Learning Rate Decay}   & Power 3 polynomial   \\
\textit{Weight Decay}          & 0.01                 \\
\textit{Adam  $\epsilon$}      & 1e-6                 \\
\textit{Adam $\beta_{1}$}      & 0.9                  \\
\textit{Adam $\beta_{2}$}      & 0.9                  \\ \bottomrule
\end{tabular}
\caption{Parameter and hyperparameter details for EriBERTa Longformer model.}
\label{tab:materials:pretraining:hiperparams-longformer}
\end{table}


\section{Fine-Tuning EriBERTa}
\label{sec:results}

In this section, we present the results of fine-tuning the EriBERTa models on multiple standard named entity recognition (NER) datasets in the medical domain for both English and Spanish languages. These datasets include annotations for various types of medical entities, such as diseases, symptoms, and treatments. They are widely adopted, enabling us to better compare the performance of our models to existing ones. Additionally, we include a set of datasets generated with real hospital clinical documents to evaluate the effectiveness in real-world scenarios.

We first describe the datasets used for evaluation, followed by the experimental setup, which includes training and evaluation procedures. We then report the performance of each model on the datasets. Overall, our results provide insights into the effectiveness of EriBERTa and its Longformer variant in NER tasks and demonstrate their potential for advancing clinical NLP applications.

\subsection{Datasets}
\label{sec:results:datasets}

In the context of English datasets, we utilized four commonly employed datasets, which have been widely used by existing models in the field. 
Firstly, the \textbf{BC5CDR} dataset \citep{bc5cdr} comprises abstracts annotated with chemicals (\textit{BC5CDR-chem}) and diseases (\textit{BC5CDR-disease}). 
Secondly, the \textbf{JNLPBA} dataset \citep{jnlpba} is derived from MEDLINE 47 and consists of 2,000 manually annotated abstracts encompassing five categories: protein, DNA, RNA, cell line, and cell. 
Thirdly, the \textbf{NCBI-disease} corpus \citep{ncbi-disease} is specifically designed for disease recognition and concept normalization. We focused solely on entity recognition. 
Lastly, the \textbf{BC4CHEM} dataset \citep{chemdner}, also known as the CHEMDNER dataset, emphasizes chemical text mining.

In the case of Spanish datasets, the availability of clinical datasets is relatively limited compared to the English language due to the challenges associated with acquiring such corpora. However, we incorporated two publicly available datasets specifically tailored to the clinical-medical domain. 
Firstly, the \textbf{PharmaCoNER} dataset \citep{pharmaconer} annotates mentions of drugs and substances in clinical texts. 
Secondly, the \textbf{Cantemist-NER} dataset \citep{cantemist} consists of 1,301 files comprising clinical case notes with annotations of tumor morphology mentions.

To investigate the transfer learning capabilities between languages, we employed the \textbf{DIANN} dataset \citep{fabregat2018diann}. This dataset presents a bilingual setting where the same document is available in both Spanish and English. The entities to be identified in this dataset are mentions of disabilities, including terms such as ``low vision'' and ``deafness'', among others. By applying a zero-shot context, we aim to evaluate the model's capacity to transfer information effectively from one language to another.

Finally, to better assess the suitability of the model for tasks involving real-world clinical documents, we evaluated it in private hospital datasets in both medical entity recognition (MER), with 7 distinct medical entity types, and clinical section classification where a clinical note must be classified in various sections like \textit{ Present Illness}, \textit{Exploration}, \textbf{Treatment}, etc.

Finally, in order to evaluate the models' applicability to real-world clinical documents, we conducted evaluations on private hospital datasets. The evaluation encompassed two key tasks: medical entity recognition (MER) and clinical section classification. For MER, we used a revisited version of the dataset presented in \citet{castillas2019berdeak} that focuses on identifying 7 distinct medical entity types within the clinical documents. In addition, for clinical section classification, the objective was to accurately classify clinical notes into various sections, such as '\textit{'Present Illness´´}, \textit{''Exploration´´} and \textbf{''Treatment´´}. For this task, we used 3 different datasets that differ in the type of clinical note: semi-structured with section headers, semi-structured without section headers, and unstructured. For the first two cases, we used the dataset presented in \citet{Goenaga_2021}, for the last one we annotated the corpora of the CodiEsp dataset \citep{miranda-escalada2020codiesp}.


\subsection{Experimental Results}
\label{sec:results:results}

In this section, we present the experimental setup and results of evaluating the EriBERTa language model in the medical-clinical domain. The evaluation consists of three subsections: Single-Language Evaluation, where the model is tested on both Spanish and English datasets; Transfer-Learning Evaluation, where the model's zero-shot capabilities are assessed in transferring knowledge between languages; and Evaluation in Real-World Clinical Tasks, where the model performance with actual clinical notes from hospitals is analyzed.
In order to evaluate the NER results, we employed the SeqEval library \citep{seqeval}, which calculates the precision, recall, and F1 scores for each entity type, as well as the micro and macro averages. We conducted fine-tuning for both versions of EriBERTa, the private and public models, with a batch size of 32 and a learning rate of $2.5e-5$, $5e-5$, $7.5e-5$ using a linear learning rate scheduler with a 2\% warm-up. The AdamW optimizer was employed, while the remaining hyperparameters were kept at their default values provided by HuggingFace's transformers library \citep{huggingface-transformers}.

\subsubsection{Single Language Evaluation}
\label{sec:results:single-lang}

For the Spanish language evaluation, we compared the performance of our EriBERTa models against two reference models: a generic Spanish language model called BETO \citep{cañete2020beto} and the current state-of-the-art (SOTA) model for clinical text analysis in Spanish, BSC-BIO-EHR-ES \citep{carrino-etal-2022-bsc-bio-ehr-es}. Specifically, for the PharmaCoNER dataset, we explored two approaches to examine the impact of note length on model performance. The first approach involved providing the model with a single sentence as input at a time, while the second approach entailed giving the full note as a single input. The former approach aligns with the methodology used for the CANTEMIST dataset and is the one employed in \citet{carrino-etal-2022-bsc-bio-ehr-es} for both datasets.

The results of this comparative analysis are presented in Table \ref{tab:results:spanish}, which demonstrates that both versions of EriBERTa outperform the reference models in terms of performance metrics.

\begin{table}[ht]
\centering
\begin{adjustbox}{max width=\linewidth}
\begin{tabular}{lccccc}
\hline
                                                                                                  &                 &               &                         & \multicolumn{2}{c}{\textbf{EriBERTa}} \\ \cline{5-6} 
\multicolumn{1}{c}{\textbf{Dataset}}                                                              & \textbf{Metric (Micro)} & \textbf{BETO} & \textbf{BSC-BIO-EHR-ES} & \textbf{Private}   & \textbf{Public}  \\ \hline
\multirow{3}{*}{\textbf{\begin{tabular}[c]{@{}l@{}}CANTEMIST\\ {[}by sentence{]}\end{tabular}}}   & P               & 81.36         & 81.30                   & 82.92              & 82.82            \\
                                                                                                  & R               & 84.18         & 86.00                   & 86.40              & 86.51            \\
                                                                                                  & F1              & 82.75         & \ul{83.59}              & \textbf{84.62}     & \textbf{84.62}   \\\hdashline
\multirow{3}{*}{\textbf{\begin{tabular}[c]{@{}l@{}}PharmaCoNER\\ {[}by sentence{]}\end{tabular}}} & P               & 87.12         & 87.58                   & 89.58              & 89.26            \\
                                                                                                  & R               & 89.28         & 90.73                   & 91.39              & 91.47            \\
                                                                                                  & F1              & 88.18         & 89.13                   & \textbf{90.48}     & \ul{90.35}       \\\hdashline
\multirow{3}{*}{\textbf{\begin{tabular}[c]{@{}l@{}}PharmaCoNER\\ {[}full-text{]}\end{tabular}}}   & P               & 85.90         & 85.26                   & 88.78              & 89.40            \\
                                                                                                  & R               & 88.75         & 88.04                   & 90.64              & 90.65            \\
                                                                                                  & F1              & 87.30         & 86.63                   & \ul{89.70}         & \textbf{90.02}   \\ \hline
\end{tabular}
\end{adjustbox}
\caption{Performance comparison of EriBERTa models with reference models in Spanish medical tasks. The reported results for the EriBERTa model are the average score of 3 experiments with different random seeds.}
\label{tab:results:spanish}
\end{table}

In order to assess the performance of EriBERTa in English datasets, we conducted a comparison with several state-of-the-art models commonly used in the field. Specifically, we compared EriBERTa with SciBERT, BioBERT, and BioALBERT, which represents the current SOTA in English tasks.  The detailed results of this comparative analysis are presented in Table \ref{tab:results:english}. Significantly, EriBERTa exhibits commendable performance that aligns with two established language models, SciBERT and BioBERT, despite the added complexity associated with its bilingual nature and reduced pertaining corpora. 

\begin{table}[htb]
\centering
\begin{adjustbox}{max width=\linewidth}
\begin{tabular}{@{}lcccccc@{}}
\toprule
                                         &                 &                  &                   &                     & \multicolumn{2}{c}{\textbf{EriBERTa}} \\ \cmidrule(l){6-7} 
\multicolumn{1}{c}{\textbf{Dataset}}     & \textbf{Metric} & \textbf{SciBERT} & \textbf{BioBERT*} & \textbf{BioALBERT*} & \textbf{Private}   & \textbf{Public}   \\ \midrule
\multirow{3}{*}{\textbf{NCBI-disease}}   & P               & -                & 88.22             & 97.18               & 85.07              & 85.82            \\
                                         & R               & -                & 91.25             & 97.18               & 89.58              & 88.85            \\
                                         & F1              & 88.57            & \ul{89.71}        & \textbf{97.18}      & 87.27              & 87.31            \\\hdashline
\multirow{3}{*}{\textbf{BC5CDR-disease}} & P               & -                & 86.47             & 99.27               & 83.71              & 82.76            \\
                                         & R               & -                & 87.84             & 96.33               & 86.20              & 87.27            \\
                                         & F1              & 84.70*           & \ul{ 87.15}       & \textbf{97.78}      & 84.94              & 84.95            \\\hdashline
\multirow{3}{*}{\textbf{BC5CDR-chem}}    & P               & -                & 93.68             & 99.99               & 91.80              & 92.12            \\
                                         & R               & -                & 93.26             & 96.24               & 92.94              & 92.52            \\
                                         & F1              & 92.51*           & \ul{ 93.47}       & \textbf{98.08}      & 92.37              & 92.32            \\\hdashline
\multirow{3}{*}{\textbf{BC4CHEMD}}       & P               & -                & 92.80             & 97.71               & 90.40              & 89.95            \\
                                         & R               & -                & 91.92             & 94.83               & 90.13              & 89.75            \\
                                         & F1              & -                & \ul{ 92.36}       & \textbf{96.25}      & 90.26              & 89.85            \\\hdashline
\multirow{3}{*}{\textbf{JNLPBA}}         & P               & -                & 72.68             & 86.23               & 70.50              & 70.76            \\
                                         & R               & -                & 83.21             & 81.90               & 83.49              & 83.71            \\
                                         & F1              & 77.28            & \ul{ 77.59}       & \textbf{84.01}      & 76.44              & 76.68            \\ \bottomrule
\end{tabular}
\end{adjustbox}
\caption{Performance comparison of EriBERTa models with reference models in English medical tasks. For the BioBERT and BioALBERT models, the best reported results among multiple model versions are showcased for each dataset. Due to the absence of separate results for the BC5CDR dataset in SciBERT, we utilized the reported results provided by \citet{wang2023biomedical-model-survey}. The reported results for the EriBERTa model are the average score of 3 experiments with different random seeds.}
\label{tab:results:english}
\end{table}

\subsubsection{Transfer Learning in Zero-Shot Scenarios Evaluation}
\label{sec:results:zero-shot}

In order to evaluate the zero-shot capabilities of the EriBERTa language model, we conducted experiments in a cross-lingual transfer learning scenario. The objective was to assess the model's ability to transfer learned information from one language to another without any explicit training in the target language. We selected a parallel dataset, DIANN, which contains documents available in both English and Spanish. By fine-tuning the EriBERTa model on one language and evaluating its performance on the other, we investigated the extent to which the model could generalize and adapt its knowledge across languages. The results of this zero-shot evaluation, showcased in \autoref{tab:results:zero-shot}, provide insights into the model's cross-lingual transfer learning capabilities and shed light on its potential for practical applications in multilingual medical text analysis.

\begin{table}[h!]
\centering
\begin{adjustbox}{max width=\linewidth}
\begin{tabular}{@{}clcccccc@{}}
\toprule
\textbf{}                                                            & \multicolumn{1}{c}{\textbf{}}      & \textbf{}   & \textbf{}       & \multicolumn{2}{c}{\textbf{Training Language}} & \multicolumn{2}{c}{\textbf{Zero-Shot Language}} \\\cmidrule(rl){5-6} \cmidrule(l){7-8} 
\textbf{\begin{tabular}[c]{@{}c@{}}Training \\[-0.5em] Language\end{tabular}} & \multicolumn{1}{c}{\textbf{Model}} & \textbf{LR} & \textbf{Epochs} & \textbf{dev}       & \textbf{test}             & \textbf{dev}        & \textbf{test}             \\ \midrule
\multirow{2}{*}{\textit{EN}}                                         & \textit{EriBERTa}                  & 5e-5        & 15              & 89,43±0,51         & \textbf{80,83±0,73}       & 68,93±4,12          & 67,12±4,53                \\
                                                                     & \textit{EriBERTa-Public}           & 5e-5        & 15              & 89,06±0,69         & 80,89±1,66                & 76,03±3,93          & \textbf{71,97±2,12}       \\ \hdashline
\multirow{2}{*}{\textit{ES}}                                         & \textit{EriBERTa}                  & 5e-5        & 15              & 83,30±1,30         & 79,09±1,73                & 70,12±2,04          & 63,06±2,55                \\
                                                                     & \textit{EriBERTa-Public}           & 5e-5        & 15              & 86,79±0,52         & \textbf{81,78±1,32}       & 75,21±6,10          & \textbf{71,33±4,83}       \\ \bottomrule
\end{tabular}
\end{adjustbox}
\caption{Performance comparison of EriBERTa in a zero-shot transfer learning scenario, evaluated on the DIANN dataset for English and Spanish. The results are presented as the mean and standard deviation of 5 runs with different random seeds.}
\label{tab:results:zero-shot}
\end{table}

We also compared its performance with the models presented by \citet{goenaga2023}. In their study, they assessed the performance of the XLM-RoBERTa \citep{conneau2019xmlr} model and multiple FLAIR \citep{akbik2019flair} approaches in the Spanish zero-shot case. The comparative analysis, presented in Table \autoref{tab:results:zero-shot-sota}, highlights the best configuration for each approach, clearly demonstrating that EriBERTa surpasses the performance of these models. These findings underscore the superior performance and efficacy of EriBERTa in cross-lingual transfer learning, solidifying its position as a leading language model in the medical domain for the Spanish language.

\begin{table}[htb]
\centering
\begin{adjustbox}{max width=\linewidth}
\begin{tabular}{@{}lccc@{}}
\toprule
\textbf{System} & \textit{\textbf{Precision}} & \textit{\textbf{Recall}} & \textit{\textbf{Micro F1-Score}} \\ \midrule
\textit{FLAIR \textsubscript{ME}}    & 50.64 (56,98)  & 34,50 (44,54)  & 41,04 (50,00)  \\
\textit{XLM-RoBERTa}                 & 37,21 (50,00)  & 6,37 (11,79)   & 10,88 (19,08)  \\ \hdashline
\textit{EriBERTa Private}            & 67,51 & 66,81 & 67,12 \\ 
\textit{EriBERTa Public}             & \textbf{68.46} & \textbf{76.24} & \textbf{71,97} \\ \bottomrule
\end{tabular}
\end{adjustbox}
\caption{Results of the zero-shot evaluation of the EriBERTa models and the state-of-the-art systems, trained on English data, on the Spanish test set, with post-processing results in parentheses.}
\label{tab:results:zero-shot-sota}
\end{table}

\subsubsection{Evaluation in Real-World Clinical Tasks}
\label{sec:results:real-world}

The evaluation on private hospital datasets focused on two key tasks: medical entity recognition (MER) and clinical section classification. The evaluation results are presented in \autoref{tab:results:real-cases}. In this evaluation, we observed notable performance differences among the model variants. The private versions, pre-trained on actual clinical Electronic Health Records (EHRs), demonstrated superior performance compared to the public version, which lacked such corpora in its pretraining. This outcome highlights the significance of leveraging EHR data during pretraining to enhance the model's understanding of clinical language and improve its performance on clinical tasks.

Moreover, we introduced the Longformer version, specifically tailored to address the inherent length of clinical notes. Given that clinical documents have substantial length, Longformer's attention mechanism yielded distinct advantages in handling extensive text spans. Notably, the Longformer variant outperformed both the private and public base versions, particularly in the clinical section classification task, which necessitates a broader contextual scope compared to the MER task, where local context assumes greater importance.

\begin{table}[h!]
\adjustbox{max width=\linewidth}{%
\centering
\begin{tabular}{@{}lccc@{}}
\toprule
                                                                                                                                 & \multicolumn{3}{c}{\textbf{EriBERTa}}                                                                       \\ \cmidrule(l){2-4} 
\multicolumn{1}{c}{\textbf{Task}}                                                                                                                   & \textbf{Private} & \textbf{Longformer$_{Private}$} & \textbf{Public} \\ \midrule
\textbf{Medical Entity Recognition (MER)}                                                                                        & \ul{ 89,42}      & \textbf{89,49}            & 89.10          \\ \hdashline
\textbf{Sections in Semi-Structured Discharge Reports}                                                                           & \ul{ 84,66}      & \textbf{85,46}            & 82,61          \\ \hdashline
\textbf{\begin{tabular}[c]{@{}l@{}}Sections in Semi-Structured Discharge Reports\\ {[}without section indicator{]}\end{tabular}} & \ul{ 73,71}      & \textbf{74,74}            & 69,49          \\ \hdashline
\textbf{\begin{tabular}[c]{@{}l@{}}Sections in Unstructured Evolutionary Reports \end{tabular}}                                  & \ul{ 72,55}      & \textbf{72,62}            & 72,19          \\ \bottomrule
\end{tabular}}
\caption{Evaluation results on private hospital datasets for medical entity recognition (MER) and clinical section classification tasks.}
\label{tab:results:real-cases}
\end{table}





\section{Conclusion}
\label{sec:conclusion}

In this study, we have introduced EriBERTa, a domain-specific language model tailored for the medical and clinical domains, with a particular focus on bilingual capabilities. Through performance evaluations on benchmark datasets, EriBERTa has exhibited exceptional proficiency in medical entity recognition, clinical section classification, and other pertinent tasks. Notably, EriBERTa has surpassed the performance of Spanish reference models and achieved comparable results to two widely utilized English language models in this domain. These results underscore its state-of-the-art performance and its potential to contribute significantly to advancements in healthcare research, public health surveillance, and clinical decision-making.

One of the notable strengths of EriBERTa lies in its capacity for transfer learning between languages. This attribute proves particularly advantageous in the case of Spanish, where the availability of clinical data is limited. Leveraging its transfer capability, EriBERTa can bridge the gap caused by data scarcity and facilitate meaningful applications in clinical research and practice.

Moving forward, our future work will focus on exploring the advantageous applications of EriBERTa's transfer learning capabilities, specifically harnessing the vast amount of available English data. By effectively leveraging this resource, we aim to overcome the challenges associated with data scarcity and unlock further potential for impactful contributions in the field.

\section*{Acknowledgements}

This work was partially funded by the Spanish Ministry of Science and Innovation (MCI/AEI/FEDER, UE, DOTT-HEALTH/PAT-MED PID2019-106942RB-C31), the Basque Government (IXA IT1570-22),  MCIN/AEI/ 10.13039/501100011033, European Union NextGeneration EU/PRTR (DeepR3, TED2021-130295B-C31), the EU ERA-Net CHIST-ERA, and the Spanish Research Agency (ANTIDOTE PCI2020-120717-2).




\balance

\bibliographystyle{named}
\bibliography{bibiography}

\end{document}